# Incorporating Side Information in Probabilistic Matrix Factorization with Gaussian Processes

Ryan Prescott Adams*  George E. Dahl
Department of Computer Science
University of Toronto
Toronto, Canada

Iain Murray
School of Informatics
University of Edinburgh
Edinburgh, Scotland

## Abstract

Probabilistic matrix factorization (PMF) is a powerful method for modeling data associated with pairwise relationships, finding use in collaborative filtering, computational biology, and document analysis, among other areas. In many domains, there are additional covariates that can assist in prediction. For example, when modeling movie ratings, we might know when the rating occurred, where the user lives, or what actors appear in the movie. It is difficult, however, to incorporate this side information into the PMF model. We propose a framework for incorporating side information by coupling together multiple PMF problems via Gaussian process priors. We replace scalar latent features with functions that vary over the covariate space. The GP priors on these functions require them to vary smoothly and share information. We apply this new method to predict the scores of professional basketball games, where side information about the venue and date of the game are relevant for the outcome.

## 1 Introduction

Many data that we wish to analyze are best modeled as the result of a pairwise interaction. The pair in question might describe an interaction between items from different sets, as in collaborative filtering, or might describe an interaction between items from the same set, as in social network link prediction. The salient feature of these *dyadic* data modeling tasks is that the observations are the result of interactions: in the popular Netflix prize example, one is given user/movie pairs with associated ratings and must predict the ratings of unseen pairs. Other examples of this sort of relational data include biological pathway analysis, document modeling, and transportation route discovery.

One approach to relational data treats the observations as a matrix and then uses a probabilistic low-rank approximation to discover structure in the data. This approach was pioneered by Hofmann (1999) to model word co-occurrences in text data. These *probabilistic matrix factorization* (PMF) models have generated a great deal of interest as powerful methods for modeling dyadic data. See Srebro (2004) for a discussion of approaches to machine learning based on matrix factorization and Salakhutdinov and Mnih (2008a) for a current view on applying PMF in practice.

One difficulty with the PMF model is that there are often more data available about the observations than simply the identities of the participants. Often the interaction itself will have additional labels that are relevant to prediction. In collaborative filtering, for example, the date of a rating is known to be important (Koren, 2009). Incorporating this side information directly as part of the low-rank feature model, however, limits the effect to only simple, linear interactions. In this paper we present a generalization of probabilistic matrix factorization that replaces scalar latent features with functions whose inputs are the side information. By placing Gaussian process priors on these latent functions, we achieve a flexible nonparametric Bayesian model that incorporates side information by introducing dependencies between PMF problems.

## 2 The Dependent PMF Model

In this section we present the *dependent probabilistic matrix factorization* (DPMF) model. The objective of DPMF is to tie together several related probabilistic matrix factorization problems and exploit side information by incorporating it into the latent features. We introduce the standard PMF model first and then show how it can be extended to include this side information.

---

*http://www.cs.toronto.edu/~rpa

## 2.1 Probabilistic Matrix Factorization

In the typical probabilistic matrix factorization framework, we have two sets, $\mathcal{M}$ and $\mathcal{N}$, of sizes $M$ and $N$. For a collaborative filtering application, $\mathcal{M}$ might be a set of films and $\mathcal{N}$ might be a set of users. $\mathcal{M}$ and $\mathcal{N}$ may also be the same sets, as in the basketball application we explore later in this paper. In general, we are interested in the outcomes of interactions between members of these two sets. Again, in the collaborative filtering case, the interaction might be a rating of film $m$ by user $n$. In our basketball application, the observations are the scores of a game between teams $m$ and $n$. Our goal is to use the observed interactions to predict unobserved interactions. This can be viewed as a *matrix completion* task: we have an $M \times N$ matrix $Z$ in which only some entries are observed and must predict some of the unavailable entries.

One approach is to use a generative model for $Z$. If this model describes useful interactions between the rows and columns of $Z$, then inference can provide predictions of the unobserved entries. A typical formulation draws $Z$ from a distribution that is parameterized by an unobserved matrix $Y$. This matrix $Y$ is of rank $K \ll M, N$ so that we may write $Y = UV^\mathsf{T}$, where $U$ and $V$ are $M \times K$ and $N \times K$ matrices, respectively. A common approach is to say that the rows of $U$ and $V$ are independent draws from two $K$-dimensional Gaussian distributions (Salakhutdinov and Mnih, 2008b). We then interpret these $K$-dimensional vectors as latent features that are distributed representations of the members of $\mathcal{M}$ and $\mathcal{N}$. We denote these vectors as $u_m$ and $v_n$ for the (transposed) $m$th row of $U$ and $n$th row of $V$, respectively, so that $Y_{m,n} = u_m^\mathsf{T} v_n$.

The distribution linking $Y$ and $Z$ is application-specific. For ratings data it may be natural to use an ordinal regression model. For binary data, such as in link prediction, a Bernoulli logistic model may be appropriate. PMF models typically assume that the entries of $Z$ are independent given $Y$, although this is not necessary. In Section 4 we will use a conditional likelihood model that explicitly includes dependencies.

## 2.2 Latent Features as Functions

We now generalize the PMF model to include side information about the interactions. Let $\mathcal{X}$ denote the space of such side information, and $x$ denote a point in $\mathcal{X}$. The time of a game or of a movie rating are good examples of such side information, but it could also involve various features of the interaction, features of the participants, or general nuisance parameters.

To enable dependence on this side information, we extend the PMF model by replacing latent

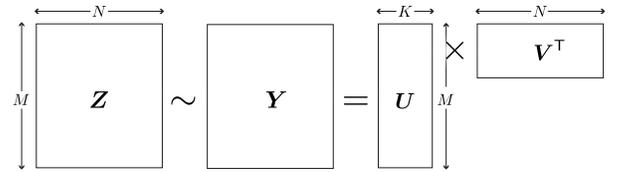

(a) Standard probabilistic matrix factorization

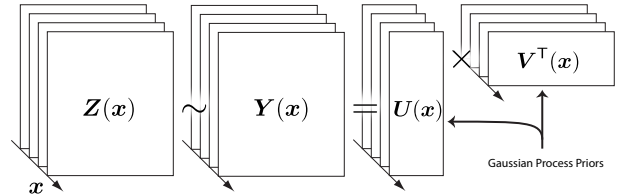

(b) Dependent probabilistic matrix factorization

Figure 1: (a) The basic low-rank matrix factorization model uses a matrix $Y$ to parameterize a distribution on random matrices, from which $Z$ (containing the observations) is taken to be a sample. The matrix $Y$ is the product of two rank-$K$ matrices $U$ and $V$. (b) In dependent probabilistic matrix factorization, we consider the low-rank matrices to be "slices" of functions over $x$ (coming out of the page). These functions have Gaussian process priors.

feature vectors $u_m$ and $v_n$, with *latent feature functions* $u_m(x) : \mathcal{X} \to \mathbb{R}^K$ and $v_n(x) : \mathcal{X} \to \mathbb{R}^K$. The matrix $Y$ is now a function $Y(x)$ such that $Y_{m,n}(x) = u_m^\mathsf{T}(x) v_n(x)$, or alternatively, $Y(x) = U(x)V^\mathsf{T}(x)$, where the $Z(x)$ matrix is drawn according to a distribution parameterized by $Y(x)$. We model each $Z(x)$ as conditionally independent, given $Y(x)$. This representation, illustrated in Figs. 1 and 2, allows the latent features to vary according to $x$, capturing the idea that the side information should be relevant to the distributed representation. We use a multi-task variant of the Gaussian process as a prior for these vector functions to construct a nonparametric Bayesian model of the latent features.

## 2.3 Multi-Task Gaussian Process Priors

When incorporating functions into Bayesian models, we often have general beliefs about the functions rather than knowledge of a specific basis. In these cases, the Gaussian process is a useful prior, allowing for the general specification of a distribution on functions from $\mathcal{X}$ to $\mathbb{R}$ via a positive-definite covariance kernel $C(x, x') : \mathcal{X} \times \mathcal{X} \to \mathbb{R}$ and a mean function $\mu(x) : \mathcal{X} \to \mathbb{R}$. For a general review of Gaussian processes for machine learning see Rasmussen and Williams (2006). In this section we restrict our discussion to GP priors for the functions $u_m(x)$, but we deal with the $v_n(x)$ functions identically.

It is reasonable for the feature function $u_m(x)$ to be independent of another individual's feature func-

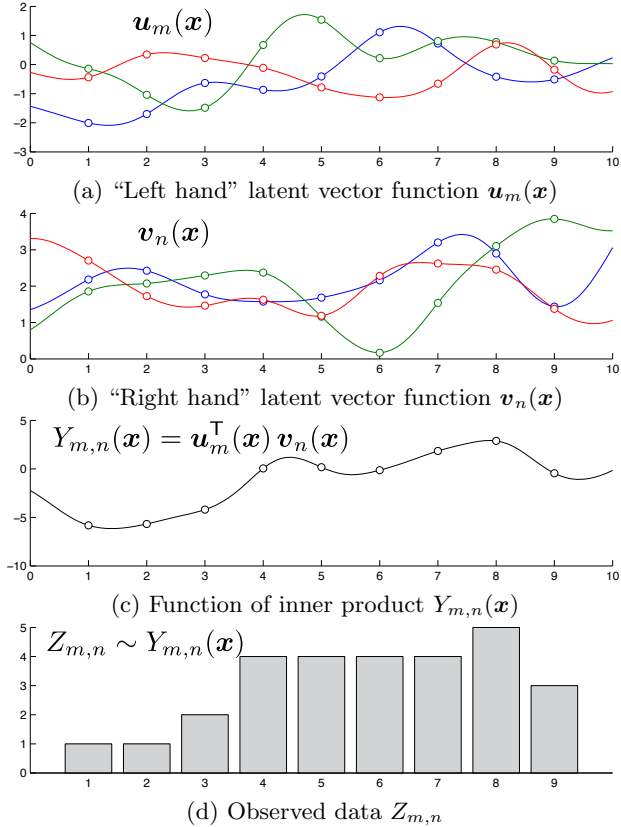

(a) "Left hand" latent vector function $\boldsymbol{u}_m(\boldsymbol{x})$

(b) "Right hand" latent vector function $\boldsymbol{v}_n(\boldsymbol{x})$

(c) Function of inner product $Y_{m,n}(\boldsymbol{x})$

(d) Observed data $Z_{m,n}$

Figure 2: These figures illustrate the generative view of the DPMF model. (a,b) Vector functions for $m$ and $n$ are drawn from from Gaussian processes. (c) The pointwise inner product of these functions yields the latent function $Y_{m,n}(\boldsymbol{x})$. (d) The observed data, in this case, ordinal values between 1 and 5, that depend on $Y_{m,n}(\boldsymbol{x})$.

tion $\boldsymbol{u}_{m'}(\boldsymbol{x})$, but we would like for each of the components *within* a particular function $\boldsymbol{u}_m(\boldsymbol{x})$ to have a structured prior. Rather than use independent Gaussian processes for each of the $K$ scalar component functions in $\boldsymbol{u}_m(\boldsymbol{x})$, we use a multi-task GP as in Teh et al. (2005) and Bonilla et al. (2008). We perform a pointwise linear transformation of $K$ independent latent functions using a matrix $L_{\Sigma_U}$ that is the Cholesky decomposition of an inter-task covariance matrix $\Sigma_U$, i.e., $\Sigma_U = L_{\Sigma_U} L_{\Sigma_U}^\mathsf{T}$. The covariance functions $C_k^U(\boldsymbol{x}, \boldsymbol{x}')$ (and hyperparameters $\theta_k^U$) are shared across the members of $\mathcal{M}$, and constant mean functions $\boldsymbol{\mu}_U(\boldsymbol{x})$ are added to each function after the linear transformation.

The *intra-feature* sharing of the covariance function, mean function, and hyperparameters is intended to capture the idea that the characteristic variations of features will tend to be consistent across members of the set. If a feature function learns to capture, for example, whether or not users in a collaborative filtering problem enjoy Christmas movies, then we might expect them to share an annual periodic variation. The *inter-feature* covariance matrix $\Sigma_U$, on the other hand, captures the idea that some features are informative about others and that this information can be shared. Salakhutdinov and Mnih (2008b) applied this idea to scalar features.

We refer to this model as *dependent* probabilistic matrix factorization because it ties together a set of PMF problems which are indexed by $\mathcal{X}$. This yields a useful spectrum of possible behaviors: as the length scales of the GP become large and the side information in $\boldsymbol{x}$ becomes uninformative, then our approach reduces to a single PMF problem; as the length scales become small and the side information becomes highly informative then each unique $\boldsymbol{x}$ has its own PMF model. The marginal distribution of a PMF problem for a given $\boldsymbol{x}$ are the same as that in Salakhutdinov and Mnih (2008b). Additionally, by having each of the $K$ feature functions use different hyperparameters, the variation over $\mathcal{X}$ can be shared differently for the features. In a sports modeling application, one feature might correspond to coaches and others to players. Player personnel may change at different timescales than coaches and this can be captured in the model.

### 2.4 Constructing Correlation Functions

One of the appeals of using a fully-Bayesian approach is that it in principle allows hierarchical inference of hyperparameters. In the DPMF case, we may not have a strong preconception as to precisely how useful the side information is for prediction. The relevance of side information can be captured by the length-scales of the covariance functions on $\mathcal{X}$ (Rasmussen and Williams, 2006). If $\mathcal{X}$ is a $D$-dimensional real space, then a standard choice is the automatic relevance determination (ARD) covariance function:

$$C_{\mathsf{ARD}}(\boldsymbol{x}, \boldsymbol{x}') = \exp\left\{-\frac{1}{2}\sum_{d=1}^D (x_d - x'_d)^2/\ell_d^2\right\}, \quad (1)$$

where in our notation there would be $2K$ sets of length scales that correspond to the covariance hyperparameters $\{\theta_k^U\}$ and $\{\theta_k^V\}$.

While the ARD prior is a popular choice, some DPMF applications may have temporal data that cause feature functions to fluctuate periodically, as in the previous Christmas movie example. For this situation it may be appropriate to include a periodic kernel such as

$$C_{\mathsf{per}}(x, x') = \exp\left\{-2\sin^2\left(\frac{1}{2}(x-x')\right)/\ell^2\right\}. \quad (2)$$

Note that we have defined both of these as *correlation* functions (unit marginal variances), and allow for variation in function amplitudes to be captured via $\Sigma_U$ and $\Sigma_V$ as in Bonilla et al. (2008).

### 2.5 Reducing Multimodality

When performing inference, overcompleteness can cause difficulties by introducing spurious modes into the posterior distribution. It is useful to construct parameterizations that avoid multimodality. Although we have developed our notation as $\boldsymbol{UV}^\mathsf{T}$, in practice we restrict the right-hand factor to be positive via a component-wise transformation $\psi(r) = \ln(1 + e^r)$ of $\boldsymbol{V}$ so that $\boldsymbol{Y}(\boldsymbol{x}) = \boldsymbol{U}(\boldsymbol{x})\psi(\boldsymbol{V}^\mathsf{T}(\boldsymbol{x}))$. In product models such as this, there are many posterior modes corresponding to sign flips in the functions (Adams and Stegle, 2008). Restricting the sign of one factor improves inference without making the model less expressive.

### 2.6 Summary of Model

For clarity, we present an end-to-end generative view of the DPMF model: 1) Two sets of $K$ GP hyperparameters, denoted $\{\theta_k^U\}$ and $\{\theta_k^V\}$, come from top-hat priors; 2) $K(M+N)$ functions are drawn from the $2K$ Gaussian processes, these are denoted below as $\{\boldsymbol{f}_{k,m}^U\}$ and $\{\boldsymbol{f}_{k,n}^V\}$; 3) Two $K$-dimensional mean vectors $\boldsymbol{\mu}_U$ and $\boldsymbol{\mu}_V$ come from vague Gaussian priors; 4) Two $K{\times}K$ cross-covariance matrices $\Sigma_U$ and $\Sigma_V$ are drawn from uninformative priors on positive definite matrices; 5) The "horizontally sliced" functions $\{\boldsymbol{f}_m^U\}$ and $\{\boldsymbol{f}_n^V\}$ are transformed with the Cholesky decomposition of the appropriate cross-covariance matrix and the mean vectors are added. 6) The transformation $\psi(\cdot)$ is applied elementwise to the resulting $\{\boldsymbol{v}_n(\boldsymbol{x})\}$ to make them strictly positive; 7) The inner product of $\boldsymbol{u}_m(\boldsymbol{x})$ and $\psi(\boldsymbol{v}_n(\boldsymbol{x}))$ computes $y_{m,n}(\boldsymbol{x})$; 8) The matrix $\boldsymbol{Y}(\boldsymbol{x})$ parameterizes a model for the entries of $\boldsymbol{Z}(\boldsymbol{x})$. Ignoring the vague priors, this is given by:

$$\boldsymbol{Z}(\boldsymbol{x}) \sim p(\boldsymbol{Z} \,|\, \boldsymbol{Y}(\boldsymbol{x})) \qquad \boldsymbol{Y}(\boldsymbol{x}) = \boldsymbol{U}(\boldsymbol{x})\,\psi(\boldsymbol{V}^\mathsf{T}(\boldsymbol{x}))$$
$$\boldsymbol{u}_m(\boldsymbol{x}) = L_{\Sigma_U}\boldsymbol{f}_m^U + \boldsymbol{\mu}_U \quad \boldsymbol{v}_n(\boldsymbol{x}) = L_{\Sigma_V}\boldsymbol{f}_n^V + \boldsymbol{\mu}_V$$
$$\boldsymbol{f}_{k,m}^U \sim \mathcal{GP}(\boldsymbol{x}_m, \theta_k^U) \qquad \boldsymbol{f}_{k,n}^V \sim \mathcal{GP}(\boldsymbol{x}_n, \theta_k^V).$$

### 2.7 Related Models

There have been several proposals for incorporating side information into probabilistic matrix factorization models, some of which have used Gaussian processes.

The Gaussian process latent variable model (GPLVM) is a nonlinear dimensionality reduction method that can be viewed as a kernelized version of PCA. Lawrence and Urtasun (2009) observes that PMF can be viewed as a particular case of PCA and use the GPLVM as a "kernel trick" on the inner products that produce $\boldsymbol{Y}$ from $\boldsymbol{UV}^\mathsf{T}$. The latent representations are optimized with stochastic gradient descent. This model differs from ours in that we use the GP to map from observed side information to the latent space, while theirs maps from the latent space into the matrix entries. Lawrence and Urtasun (2009) also mention the use of movie-specific metadata to augment the latent space in their collaborative filtering application. We additionally note that our model allows arbitrary link functions between the latent matrix $\boldsymbol{Y}$ and the observations $\boldsymbol{Z}$, including dependent distributions, as discussed in Section 4.2.

Another closely-related factorization model is the stochastic relational model (SRM) (Yu et al., 2007). Rather than representing $\mathcal{M}$ and $\mathcal{N}$ as finite sets, the SRM uses arbitrary spaces as index sets. The GP provides a distribution over maps from this "identity space" to the latent feature space. The SRM differs from the DPMF in that the input space for our Gaussian process corresponds to the observations themselves, and not just to the participants in the relation. Additionally, we allow each member of $\mathcal{M}$ and $\mathcal{N}$ to have $K$ functions, each with a different GP prior that may have different dependencies on $\mathcal{X}$.

A potential advantage of the DPMF model we present here, relative to the GPLVM and SRM, is that the GP priors need only be defined on the data associated with the observations for a single individual. As inference in Gaussian processes has cubic computational complexity, it is preferable to have more independent GPs with fewer data in each one than a few GPs that are each defined on many thousands of input points.

There has also been work on explicitly incorporating temporal information into the collaborative filtering problem, most notably by the winner of the Netflix prize. Koren (2009) included a simple drift model for the latent user features and baseline ratings. When rolled into the SVD learning method, this temporal information significantly improved predictive accuracy.

## 3 MCMC Inference and Prediction

As discussed in Section 2.1, the typical objective when using probabilistic matrix factorization is to predict unobserved entries in the matrix. For the DPMF model, as in the Bayesian PMF model (Salakhutdinov and Mnih, 2008a), inference and prediction are not possible in closed form. We can use Markov chain Monte Carlo (MCMC), however, to sample from the posterior distribution of the various parameters and latent variables in the model. We can then use these samples to construct a Monte Carlo estimate of the predictive distribution. If the entries of interest in $\boldsymbol{Z}(\boldsymbol{x})$ can be easily sampled given $\boldsymbol{Y}(\boldsymbol{x})$ — as is typically the case — then samples from the posterior on $\boldsymbol{Y}(\boldsymbol{x})$ allow us to straightforwardly generate predictive samples, which are the quantities of interest, and integrate out all of the latent variables.

In the DPMF model, we define the state of the Markov chain with: 1) the values of the latent feature functions $U(x)$ and $V(x)$, evaluated at the observations; 2) the hyperparameters $\{\theta_k^U, \theta_k^V\}_{k=1}^K$ associated with the feature-wise covariance functions, typically capturing the relevance of the side information to the latent features; 3) the feature cross-covariances $\Sigma_U$ and $\Sigma_V$; 4) the feature function means $\mu_U$ and $\mu_V$; 5) any parameters controlling the conditional likelihood of $Z(x)$ given $Y(x)$. Note that due to the convenient marginalization properties of the Gaussian process, it is only necessary to represent the values of feature functions at places (in the space of side information) where there have been observations.

## 3.1 Slice Sampling

When performing approximate inference via Markov chain Monte Carlo, one constructs a transition operator on the state space that leaves the posterior distribution invariant. The transition operator is used to simulate a Markov chain. Under mild conditions, the distribution over resulting states evolves to be closer and closer to the true posterior distribution (e.g., Neal (1993)). While a generic operator, such as Metropolis–Hastings or Hamiltonian Monte Carlo, can be implemented, we seek efficient methods that do not require extensive tuning. To that end, we use MCMC methods based on *slice sampling* (Neal, 2003) when performing inference in the DPMF model. Some of the variables and parameters required special treatment, detailed in the next two subsections, for slice sampling to work well.

## 3.2 Elliptical Slice Sampling

Sampling from the posterior distribution over latent functions with Gaussian process priors is often a difficult task and can be slow to mix, due to the structure imposed by the GP prior. In this case, we have several collections of functions in $U(x)$ and $V(x)$ that do not lend themselves easily to typical methods such as Gibbs sampling. Recently, a method has been developed to specifically enable efficient slice sampling of complicated Gaussian process models with no tuning or gradients (Murray et al., 2010). This method, called *elliptical slice sampling* (ESS), takes advantage of invariances in the Gaussian distribution to make transitions that are never vetoed by the highly-structured GP prior, even when there are a large number of such functions as in the DPMF.

## 3.3 Sampling GP Hyperparameters

As discussed in Section 2.4, the length scales in the covariance (correlation) functions of the Gaussian processes play a critical role in the DPMF model. It is through these hyperparameters that the model weighs the effect of side information on the predictions. In the DPMF model, a set of hyperparameters $\theta_k^U$ (or $\theta_k^V$) affect $M$ (or $N$) functions. The typical approach to this would be to fix the relevant functions $\{f_{k,m}^U\}_{m=1}^M$ and sample from the conditional posterior:

$$p(\theta_k^U \mid \{f_{k,m}^U\}_{m=1}^M) \propto p(\theta_k^U) \prod_{m=1}^M \mathcal{N}(f_{k,m}^U; 0, \Xi_{k,m}^U),$$

where $\Xi_{k,m}^U$ is the matrix that results from applying the covariance function with hyperparameters $\theta_k^U$ to the set of side information for $m$. In practice, the Markov chain on this distribution can mix very slowly, due to the strong constraints arising from the $M$ functions, despite the relative weakness of the conditional likelihood on the data. We therefore use an approach similar to Christensen and Waagepetersen (2002), which mixes faster in our application, but still leaves the posterior distribution on the hyperparameters invariant.

The model contains several draws from a Gaussian process of the form: $f_{k,n}^V \sim \mathcal{GP}(x_n, \theta_k^V)$. Consider a vector of evaluations of one of these latent functions that is marginally distributed as $f \sim \mathcal{N}(m, \Xi_\theta)$. Under the generative process, the distribution over the latent values is strongly dependent on the hyperparameters $\theta$ that specify the covariance. As a result, the posterior conditional distribution over the hyperparameters for fixed latent values will be strongly peaked, leading to slow mixing of a Markov chain that updates $\theta$ for fixed $f$. Several authors have found it useful to reparameterize Gaussian models so that under the prior the latent values are independent of each other and the hyperparameters. This can be achieved by setting $\nu = L_\theta^{-1}(f - m)$, where $L_\theta$ is a matrix square root, such as the Cholesky decomposition, of the covariance $\Xi_\theta$. Under the new prior representation, $\nu$ is drawn from a spherical unit Gaussian for all $\theta$.

We slice sample the GP hyperparameters after reparameterizing all vectors of latent function evaluations. As the hyperparameters change, the function values $f = m + L_\theta \nu$ will also change to satisfy the covariance structure of the new settings. Having observed data, some $f$ settings are very unlikely; in the reparameterized model the likelihood terms will restrict how much the hyperparameters can change. In the application we consider, with very noisy data, these updates work much better than updating the hyperparameters for fixed $f$. In problems where the data strongly restrict the possible changes in $f$, more advanced reparameterizations are possible (Christensen et al., 2006). We have developed related slice sampling methods that are easy to apply (Murray and Adams, 2010).

## 4 DPMF for Basketball Outcomes

To demonstrate the utility of the DPMF approach, we use our method to model the scores of games in the National Basketball Association (NBA) in the years 2002 to 2009. This task is appealing for several reasons: 1) it is medium-sized, with about ten thousand observations; 2) it provides a natural censored-data evaluation setup via a "rolling predictions" problem; 3) expert human predictions are available via betting lines; 4) the properties of teams vary over time as players are traded, retire and are injured; 5) other side information, such as which team is playing at home, is clearly relevant to game outcomes. Using basketball as a testbed for probabilistic models is not a new idea. In the statistics literature there have been previous studies of collegiate basketball outcomes by Schwertman et al. (1991), Schwertman et al. (1996), and Carlin (1996), although with smaller data sets and narrower models.

We use the DPMF to model the scores of games. The observations in the matrix $Z(x)$ are the actual scores of the games with side information $x$. $Z_{m,n}(x)$ is the number of points scored by team $m$ against team $n$, and $Z_{n,m}(x)$ is the number of points scored by team $n$ against team $m$. We model these with a bivariate Gaussian distribution, making this a somewhat unusual PMF-type model in that we see *two* matrix entries with each observation and we place a joint distribution over them. We use a single variance for all observations and allow for a correlation between the scores of the two teams. While the Gaussian model is not a perfect match for the data — scores are non-negative integers – each team in the NBA tends to score about 100 points per game, with a standard deviation of about ten so that very little mass is assigned to negative numbers.

Even though both sets in this dyadic problem are the same, i.e., $\mathcal{M} = \mathcal{N}$, we use different latent feature functions for $U(x)$ and $V(x)$. This makes the $U(x)$ functions of offense and the $V(x)$ functions of defense, contributing to "points for" and "points against", respectively. This specialization allows the GP hyperparameters to differ between offense and defense, so that side information affects the number of points scored and conceded in different ways.

### 4.1 Problem Setup

To use NBA basketball score prediction as a task to determine the value that using side information in our framework provides relative to the standard PMF model, we set up a rolling censored-data problem. We divided each of the eight seasons into four-week blocks. For each four-week block, the models were asked to make predictions about the games during that interval using only information from the past. In other words, when making predictions for the month of February 2005, the model could only train on data from 2002 through January 2005. We rolled over each of these intervals over the entire data set, retraining the model each time and evaluating the predictions. We used three metrics for evaluation: 1) mean predictive log probability from a Rao–Blackwellized estimator; 2) error in the binary winner-prediction task; 3) root mean-squared error (RMSE) of the two-dimensional score vector.

The winner accuracies and RMSE can be compared against human experts, as determined by the betting lines associated with the games. Sports bookmakers assign in advance two numbers to each game, the *spread* and the *over/under*. The spread is a number that is added to the score of a specified team to yield a bettor an even-odds return. For example, if the spread between the LA Lakers and the Cleveland Cavaliers is "-4.5 for the Lakers" then a single-unit bet for the Lakers yields a single-unit return if the Lakers win by 4.5 points or more (the half-point prevents ties, or *pushes*). If the Lakers lose or beat the Cavaliers by fewer than 4.5 points, then a single-unit bet on the Cavaliers would win a single-unit return. The over/under determines the threshold of a single-unit bet on the sum of the two scores. For example, if the over/under is 210.5 and the final score is 108 to 105, then a bettor who "took the over" with a single-unit bet would win a single-unit return, while a score of 99 to 103 would cause a loss (or a win to a bettor who "took the under").

From the point of view of model evaluation, these are excellent predictions, as the spread and over/under themselves are set by the bookmakers to balance bets on each side. This means that expert humans exploit any data available (e.g., referee identities and injury reports, which are not available to our model) to exert market forces that refine the lines to high accuracy. The sign of the spread indicates the favorite to win. To determine the implied score predictions themselves, we can solve a simple linear system:

$$\begin{bmatrix} 1 & 1 \\ 1 & -1 \end{bmatrix} \begin{bmatrix} \text{away score} \\ \text{home score} \end{bmatrix} = \begin{bmatrix} \text{over/under} \\ \text{home spread} \end{bmatrix}. \quad (3)$$

### 4.2 Basketball-Specific Model Aspects

As mentioned previously, the conditional likelihood function that parameterizes the distribution over the entries in $Z(x)$ in terms of $Y(x)$ is problem specific. In this application, we use

$$\begin{bmatrix} Z_{m,n}(x) \\ Z_{n,m}(x) \end{bmatrix} \sim \mathcal{N}\left( \begin{bmatrix} Y_{m,n}(x) \\ Y_{n,m}(x) \end{bmatrix}, \begin{bmatrix} \sigma^2 & \rho\sigma^2 \\ \rho\sigma^2 & \sigma^2 \end{bmatrix} \right), \quad (4)$$

where $\sigma \in \mathbb{R}^+$ and $\rho \in (-1,1)$ parameterize the bivariate Gaussian on scores and are included as part of inference. (A typical value for the correlation coefficient was $\rho=0.4$.) This allows us to easily compute the predictive log probabilities of the censored test data using a Rao–Blackwellized estimator. To do this, we sample and store predictive state from the Markov chain to construct a Gaussian mixture model. Given the predictive samples of the latent function at the new time, we can compute the means for the distribution in Equation 4. The covariance parameters are also being sampled and this forms one component in a mixture model with equal component weights. Over many samples, we form a good predictive estimate.

### 4.3 Nonstationary Covariance

In the DPMF models incorporating temporal information, we are attempting to capture fluctuations in the latent features due to personnel changes, etc. One unusual aspect of this particular application is that we expect the notion of time scale to vary depending on whether it is the off-season. The timescale appropriate during the season is almost certainly inappropriate to describe the variation during the 28 weeks between the end of one regular season and the start of another. To handle this nonstationarity of the data, we introduced an additional parameter that is the effective number of weeks between seasons, which we expect to be smaller than the true number of weeks. We include this as a hyperparameter in the covariance functions and include it as part of inference, using the same slice sampling technique described in Section 3.3. A histogram of inferred gaps for $K=4$ is shown in Figure 3. Note that most of the mass is below the true number of weeks.

### 4.4 Experimental Setup and Results

We compared several different model variants to evaluate the utility of side information. We implemented the standard fully-Bayesian PMF model using the same likelihood as above, generating predictive log probabilities as in the DPMF. We constructed DPMFs with temporal information, binary home/away information, and both of these together. We applied each of these models using different numbers of latent features, $K$, from one to five. We ran ten separate Markov chains to predict each censored interval. Within a single year, we initialized the Markov state from the ending state of the previous chain, for a "warm start". The "cold start" at the beginning of the year ran for 1000 burnin iterations, while warm starts ran for 100 iterations in each of the ten chains. After burning in and thinning by a factor of four, 100 samples of each predictive score were kept from each chain, resulting in 1000 components in the predictive Gaussian mixture model.

To prevent the standard PMF model from being too heavily influenced by older data, we only provided data to it from the current season and the previous two seasons. To prevent an advantage for the DPMF, we also limited its data in the same way. Sampling from the covariance hyperparameters in the model is relatively expensive, due to the need to compute multiple Cholesky decompositions. To improve efficiency in this regard, we ran an extensive Markov chain to burn in these hyperparameters and then fixed them for all further sampling. Without hyperparameter sampling, the remainder of the MCMC state can be iterated in approximately three minutes on a single core of a modern workstation. We performed this burn-in of hyperparameters on a span of games from 2002 to 2004 and ultimately used the learned parameters for prediction, so there is a mild amount of "cheating" on 2004 and before as those data have technically been seen already. We believe this effect is very small, however, as the covariance hyperparameters (the only state carried over) are only loosely connected to the data.

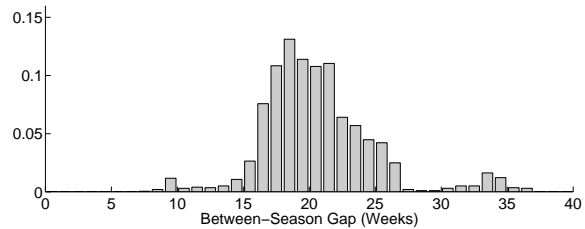

Figure 3: Histogram of the gap between seasons.

Results from these evaluations are provided in Table 1. The DPMF model demonstrated notable improvements over the baseline Bayesian PMF model, and the inclusion of more information improved predictions over the time and home/away information alone. The predictions in 2009 are less consistent, which we attribute to variance in evaluation estimates from fewer intervals being available as the season is still in progress. The effect of the number of latent features $K$ in the complex model is much less clear. Figure 4 shows the joint predictions for four different games between the Cleveland Cavaliers and the Los Angeles Lakers, using $K=3$ with time and home/away available. The differences between the predictions illustrate that the model is incorporating information both from home advantage and variation over time. The code and data are available at http://www.cs.toronto.edu/~rpa/dpmf.

## 5 Discussion

In this paper we have presented a nonparametric Bayesian variant of probabilistic matrix factorization that induces dependencies between observations via Gaussian processes. This model has the convenient property that, conditioned on the side information, the marginal distributions are equivalent to those in well-studied existing PMF models. While Gaussian processes and MCMC often carry a significant computational cost, we have developed a framework that can make useful predictions on real problems in a practical amount of time — hours for most of the predictions in the basketball problem we have studied.

There are several interesting ways in which this work could be extended. One notable issue that we have overlooked and would be relevant for many applications is that the Gaussian processes as specified in Section 2.4 only allow for smooth variation in latent features. This slow variation may be inappropriate for many models: if a star NBA player has a season-ending injury, we would expect that to be reflected better with a changepoint model (see, e.g., Barry and Hartigan (1993)) than a GP model. Also, we have not addressed the issue of how to select the number of latent features, $K$, or how to sample from this parameter. Nonparametric Bayesian models such as those proposed by Meeds et al. (2007) may give insight into this problem. Finally, we should note that other authors have explored other kinds of structured latent factors (e.g., Sutskever et al. (2009)), and there may be interesting ways to combine the features of these approaches with the DPMF.

### Acknowledgements

The authors thank Amit Gruber, Geoffrey Hinton and Richard Zemel for valuable discussions. The idea of placing basketball scores directly into the matrix was originally suggested by Daniel Tarlow. RPA is a junior fellow of the Canadian Institute for Advanced Research.

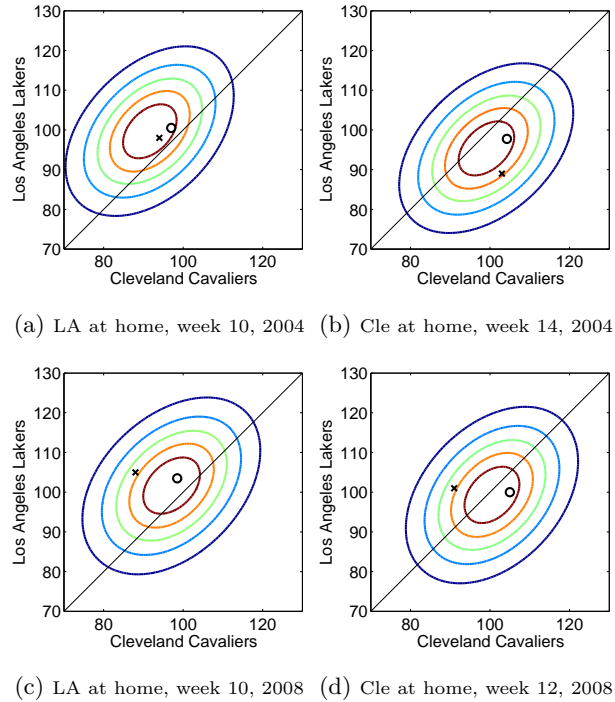

(a) LA at home, week 10, 2004  (b) Cle at home, week 14, 2004

(c) LA at home, week 10, 2008  (d) Cle at home, week 12, 2008

Figure 4: Contour plots showing the predictive densities for four games between the Cleveland Cavaliers and the Los Angeles Lakers, using $K = 3$ with home/away and temporal information available. Los Angeles is the home team in (a) and (c). Cleveland is the home team in (b) and (d). Figures (a) and (b) were in the 2004 season, (c) and (d) were in the 2008 season. The "o" shows the expert-predicted score and the "x" shows the true outcome. The diagonal line indicates the winner threshold. Note the substantial differences between home and away, even when the times are close to each other.

Table 1: Evaluations of PMF and DPMF algorithms with various numbers of latent factors. The PMF model is the fully-Bayesian approach of Salakhutdinov and Mnih (2008a), with our application-specific likelihood. DPMF(t) is the DPMF with only temporal information, DPMF(h) has only binary home/away indicators, DPMF(t,h) has both temporal and home/away information. (a) Mean predictive log probabilities of test data. (b) Error rates of winner prediction are on the left, RMSEs of scores are on the right. Expert human predictions are shown on the bottom.

|  |  | 2002 | 2003 | 2004 | 2005 | 2006 | 2007 | 2008 | 2009 | All |
|---|---|---|---|---|---|---|---|---|---|---|
| PMF | $K1$ | -7.644 | -7.587 | -7.649 | -7.580 | -7.699 | -7.733 | -7.634 | -7.653 | -7.647 |
|  | $K2$ | -7.623 | -7.587 | -7.654 | -7.581 | -7.704 | -7.738 | -7.638 | -7.673 | -7.650 |
|  | $K3$ | -7.615 | -7.586 | -7.652 | -7.581 | -7.698 | -7.734 | -7.637 | -7.666 | -7.646 |
|  | $K4$ | -7.619 | -7.585 | -7.653 | -7.581 | -7.703 | -7.734 | -7.635 | -7.667 | -7.647 |
|  | $K5$ | -7.641 | -7.589 | -7.653 | -7.580 | -7.700 | -7.736 | -7.638 | -7.667 | -7.650 |
| DPMF(t) | $K1$ | -7.652 | -7.535 | -7.556 | -7.559 | -7.665 | -7.665 | -7.618 | -7.703 | -7.620 |
|  | $K2$ | -7.620 | -7.551 | -7.580 | -7.544 | -7.675 | -7.658 | -7.611 | -7.724 | -7.621 |
|  | $K3$ | -7.620 | -7.560 | -7.605 | -7.549 | -7.673 | -7.669 | -7.611 | -7.635 | -7.615 |
|  | $K4$ | -7.618 | -7.549 | -7.585 | -7.548 | -7.673 | -7.670 | -7.608 | -7.651 | -7.613 |
|  | $K5$ | -7.640 | -7.558 | -7.591 | -7.554 | -7.669 | -7.670 | -7.609 | -7.651 | -7.618 |
| DPMF(h) | $K1$ | -7.639 | -7.549 | -7.624 | -7.549 | -7.670 | -7.706 | -7.606 | -7.627 | -7.621 |
|  | $K2$ | -7.587 | -7.553 | -7.626 | -7.551 | -7.670 | -7.707 | -7.613 | -7.640 | -7.618 |
|  | $K3$ | -7.580 | -7.542 | -7.618 | -7.539 | -7.667 | -7.706 | -7.602 | -7.637 | -7.612 |
|  | $K4$ | -7.587 | -7.545 | -7.623 | -7.547 | -7.673 | -7.704 | -7.612 | -7.652 | -7.618 |
|  | $K5$ | -7.594 | -7.541 | -7.619 | -7.544 | -7.669 | -7.709 | -7.606 | -7.643 | -7.616 |
| DPMF(t,h) | $K1$ | -7.656 | -7.515 | -7.562 | -7.534 | -7.659 | -7.662 | -7.602 | -7.670 | -7.607 |
|  | $K2$ | -7.585 | -7.515 | -7.560 | -7.520 | -7.646 | **-7.639** | -7.591 | -7.695 | -7.594 |
|  | $K3$ | **-7.579** | -7.516 | -7.563 | -7.524 | -7.651 | -7.643 | **-7.575** | **-7.586** | **-7.580** |
|  | $K4$ | -7.584 | **-7.511** | **-7.546** | -7.520 | -7.640 | -7.643 | -7.582 | -7.620 | -7.581 |
|  | $K5$ | -7.593 | -7.515 | -7.569 | **-7.512** | **-7.634** | -7.640 | -7.589 | -7.637 | -7.586 |

(a) Mean log probabilities for rolling score prediction

|  |  | 2002 | 2003 | 2004 | 2005 | 2006 | 2007 | 2008 | 2009 | All | 2002 | 2003 | 2004 | 2005 | 2006 | 2007 | 2008 | 2009 | All |
|---|---|---|---|---|---|---|---|---|---|---|---|---|---|---|---|---|---|---|---|
| PMF | $K1$ | 38.9 | 39.4 | 41.8 | 37.6 | 41.0 | 38.2 | 36.7 | 36.7 | 38.8 | 16.66 | 15.92 | 16.80 | 15.84 | 17.16 | 17.39 | 16.38 | 16.86 | 16.63 |
|  | $K2$ | 37.8 | 38.5 | 42.1 | 37.1 | 40.7 | 37.9 | 36.4 | 38.6 | 38.6 | 16.38 | 15.91 | 16.82 | 15.85 | 17.16 | 17.41 | 16.33 | 16.96 | 16.61 |
|  | $K3$ | 37.2 | 38.7 | 42.4 | 37.0 | 40.5 | 37.5 | 36.8 | 38.1 | 38.5 | 16.35 | 15.89 | 16.81 | 15.85 | 17.12 | 17.38 | 16.34 | 16.92 | 16.59 |
|  | $K4$ | 37.5 | 38.2 | 41.7 | 36.7 | 40.3 | 37.8 | 36.1 | 37.6 | 38.2 | 16.34 | 15.90 | 16.81 | 15.84 | 17.15 | 17.39 | 16.34 | 16.93 | 16.59 |
|  | $K5$ | 39.1 | 38.1 | 41.4 | 37.1 | 41.0 | 37.8 | 36.1 | 38.6 | 38.6 | 16.41 | 15.93 | 16.81 | 15.83 | 17.14 | 17.40 | 16.35 | 16.93 | 16.61 |
| DPMF(t) | $K1$ | 37.8 | 37.7 | 37.9 | 37.3 | 39.1 | 34.4 | 34.4 | 35.7 | 36.8 | 16.73 | 15.68 | 16.07 | 15.54 | 16.73 | 16.93 | 16.42 | 17.46 | 16.46 |
|  | $K2$ | 37.4 | 38.0 | 37.5 | 39.2 | 40.3 | 34.3 | 33.6 | 37.6 | 37.3 | 16.37 | 15.70 | 16.30 | 15.29 | 16.84 | 16.75 | 16.21 | 17.57 | 16.39 |
|  | $K3$ | 37.2 | 38.9 | 39.0 | 36.5 | 38.1 | 36.3 | 34.0 | 32.4 | 36.5 | 16.39 | 15.77 | 16.49 | 15.50 | 16.97 | 16.86 | 16.05 | **16.67** | 16.34 |
|  | $K4$ | 37.7 | 38.6 | 37.4 | 36.0 | 38.8 | 35.7 | 34.6 | 35.7 | 36.8 | 16.37 | 15.63 | 16.30 | 15.46 | 16.90 | 16.90 | 16.11 | 16.91 | 16.33 |
|  | $K5$ | 38.6 | 39.4 | 38.0 | 35.9 | 38.4 | 37.0 | 34.1 | 36.7 | 37.2 | 16.39 | 15.71 | 16.37 | 15.48 | 16.90 | 16.85 | 16.16 | 16.84 | 16.35 |
| DPMF(h) | $K1$ | 37.3 | 37.7 | 39.3 | 34.3 | 38.1 | 37.9 | 34.3 | 29.5 | 36.1 | 16.62 | 15.88 | 16.40 | 15.59 | 16.87 | 17.01 | 16.14 | 16.70 | 16.41 |
|  | $K2$ | 36.8 | 37.1 | 38.4 | 34.3 | 38.3 | 37.6 | 34.7 | 31.0 | 36.0 | 16.11 | 15.93 | 16.44 | 15.60 | 16.88 | 17.04 | 16.15 | 16.79 | 16.37 |
|  | $K3$ | 36.8 | 37.4 | 38.4 | 34.9 | 36.8 | 38.6 | 34.1 | 30.5 | 35.9 | 15.91 | 15.92 | 16.25 | 15.42 | 16.81 | 16.87 | 16.05 | 16.73 | 16.25 |
|  | $K4$ | 36.6 | 37.8 | 38.4 | 35.1 | 37.9 | 38.1 | 34.6 | 31.4 | 36.2 | 15.92 | 15.88 | 16.35 | 15.51 | 16.88 | 16.98 | 16.13 | 16.90 | 16.33 |
|  | $K5$ | **35.1** | 36.9 | 38.5 | 34.5 | 37.4 | 38.0 | 34.1 | 30.0 | 35.6 | 16.08 | 15.89 | 16.28 | 15.50 | 16.85 | 17.05 | 16.08 | 16.77 | 16.32 |
| DPMF(t,h) | $K1$ | 37.2 | 36.1 | 36.7 | 35.3 | 38.6 | 33.7 | 32.0 | 37.1 | 35.8 | 16.76 | **15.55** | 16.07 | 15.46 | 16.69 | 16.91 | 16.26 | 17.26 | 16.38 |
|  | $K2$ | 36.1 | 37.0 | 36.5 | 34.4 | 38.8 | 33.6 | 32.5 | 36.7 | 35.8 | 16.08 | 15.58 | 16.04 | **15.19** | 16.59 | 16.61 | 16.07 | 17.25 | 16.19 |
|  | $K3$ | 37.0 | **34.9** | 34.4 | 33.8 | 37.0 | 34.2 | 31.7 | 30.0 | 34.1 | **15.90** | 15.72 | 15.89 | 15.35 | 16.69 | **16.57** | **15.86** | 16.71 | 16.10 |
|  | $K4$ | 36.0 | 35.2 | 34.1 | 35.0 | 37.4 | **33.2** | **31.5** | 30.5 | 34.1 | 15.93 | 15.62 | **15.75** | 15.20 | **16.52** | 16.60 | 16.03 | 16.81 | **16.07** |
|  | $K5$ | 35.3 | 35.6 | **34.0** | **33.2** | **36.2** | 33.5 | 31.7 | 32.9 | **34.0** | 16.05 | 15.66 | 15.96 | 15.21 | 16.57 | 16.60 | 16.00 | 16.97 | 16.14 |
| Expert |  | 30.9 | 32.2 | 29.7 | 31.4 | 33.3 | 30.6 | 29.8 | 29.4 | 30.9 | 14.91 | 14.41 | 14.55 | 14.70 | 15.36 | 15.18 | 14.95 | 15.49 | 14.95 |

(b) Left: percentage error on rolling winner prediction, Right: RMSE on rolling score prediction